\documentclass[conference]{IEEEtran}
\IEEEoverridecommandlockouts
\usepackage{cite}
\usepackage{amsmath,amssymb,amsfonts}
\usepackage{algorithmic}
\usepackage{mathrsfs}
\usepackage{graphicx}
\usepackage{textcomp}
\usepackage{xcolor}
\def\BibTeX{{\rm B\kern-.05em{\sc i\kern-.025em b}\kern-.08em
    T\kern-.1667em\lower.7ex\hbox{E}\kern-.125emX}}
\begin{document}

\title{MSD-LLM: Predicting Ship Detention in Port State Control Inspections with Large Language Model\\
\thanks{* corresponding author}}

\author{\IEEEauthorblockN{1\textsuperscript{st} Jiongchao Jin}
\IEEEauthorblockA{\textit{Institute of High Performance Computing} \\
\textit{Agency for Science, Technology and Research}\\
Singapore \\
jin\_jiongchao@ihpc.a-star.edu.sg }
\and
\IEEEauthorblockN{2\textsuperscript{nd} Xiuju Fu}
\IEEEauthorblockA{\textit{Institute of High Performance Computing} \\
\textit{Agency for Science, Technology and Research}\\
Singapore \\
fuxj@ihpc.a-star.edu.sg}
\and
\IEEEauthorblockN{3\textsuperscript{rd} Xiaowei Gao}
\IEEEauthorblockA{\textit{\small{Department of Earth Science and Engineering}} \\
\textit{Imperial College London}\\
London, the United Kingdom \\
x.gao24@imperial.ac.uk}
\and
\IEEEauthorblockN{4\textsuperscript{th} Tao Cheng}
\IEEEauthorblockA{\textit{\small{SpaceTimeLab, Department of Civil,}} \\
\textit{\small{Environmental and Geomatic Engineering}}\\
\textit{University College London}\\
London, the United Kingdom \\
tao.cheng@ucl.ac.uk}
\and

\IEEEauthorblockN{5\textsuperscript{th} Ran Yan*}
\IEEEauthorblockA{\textit{\small{School of Civil and Environmental Engineering}} \\
\textit{ Nanyang Technological University}\\
Singapore \\
ran.yan@ntu.edu.sg}
}
\vspace{-10em}
\maketitle

\begin{abstract}
Maritime transportation is the backbone of global trade, making ship inspection essential for ensuring maritime safety and environmental protection. Port State Control (PSC), conducted by national ports, enforces compliance with safety regulations, with ship detention being the most severe consequence, impacting both ship schedules and company reputations. Traditional machine learning methods for ship detention prediction are limited by the capacity of representation learning and thus suffer from low accuracy. Meanwhile, autoencoder-based deep learning approaches face challenges due to the severe data imbalance in learning historical PSC detention records. To address these limitations, we propose Maritime Ship Detention with Large Language Models (MSD-LLM), integrating a dual robust subspace recovery (DSR) layer-based autoencoder with a progressive learning pipeline to handle imbalanced data and extract meaningful PSC representations. Then, a large language model groups and ranks features to identify likely detention cases, enabling dynamic thresholding for flexible detention predictions. Extensive evaluations on 31,707 PSC inspection records from the Asia-Pacific region show that MSD-LLM outperforms state-of-the-art methods more than 12\% on Area Under the Curve (AUC) for Singapore ports. Additionally, it demonstrates robustness to real-world challenges, making it adaptable to diverse maritime risk assessment scenarios.
\end{abstract}


\section{Introduction}
Maritime transportation is the backbone of global trade, with over 80\% of goods by volume transported through international shipping \cite{MT2023}. The International Convention for the Safety of Life at Sea (SOLAS), established and implemented by the International Maritime Organization (IMO), is the most significant international treaty governing the safety of merchant ships at sea \cite{IMO_SOLAS}. Ship detention occurring in port state control (PSC) at national ports over the world is an enforcement action taken by port states when a vessel or its crew fails to meet the standards set by international conventions and regulations that pose risks to safety, security, or environmental protection \cite{IMO_PSCMANUAL}.

Data-driven models based on machine learning (ML) or deep learning (DL) have demonstrated effectiveness, utilizing ship particulars, specifications, and historical inspection records \cite{61wang2019development,80yan2022ship,90wu2022ship,132yang2024data}. However, ML-based models can be more insensitive to the imbalanced data distribution, failing to extract the deep interrelation among the multiple discrete ship particulars and inspection records. Meanwhile, auto-encoder-based DL models can capture more complex information but rely too much on data itself and suffer from the severe imbalanced data distribution, especially in detention, where detained ships account for less than ~5\% of all vessels\cite{NIR2024Toyko}. 

To preserve the robust deep representation learning capability of autoencoders while mitigating sensitivity to data distribution, we propose the Maritime Ship Detention Large Language Model (MSD-LLM), a novel auto-encoder and LLM-based structure to deal with practical ship detention tasks with severe data imbalance. MSD-LLM consists of two key modules: an auto-encoder representation learning module and an LLM-based prediction module. In the representation learning module, a dual subspace recovery (DSR) layer, an extension of the single subspace recovery (RSR) layer \cite{lai2019robust}, is designed to extract latent features from PSC data. The DSR layer-based model widens the gap between detained and non-detained ships by aggregating margin loss and reconstruction loss, thereby enhancing the LLM module's reasoning ability. Additionally, a progressive learning strategy in the auto-encoder module ensures adaptability to diverse data distributions under practical, varied port scenarios.

Moreover, with the recent advancements in the strong reasoning and prediction ability of LLMs, many practical tasks have adopted them as a feasible tool to make sound inferences under noise and challenging data. Considering the challenges of ship detention tasks, we employ a multi-modal LLM (M-LLM) to boost the performance of detention prediction after extracting the DSR layer-based representation from auto-encoder learning module. Unlike traditional M-LLM with a Vision Transformer (ViT) to handle image representations, in MSD-LLM, we replace the ViT with our proposed DSR layer-based representation learning module, which retains the same format and representation understanding ability of M-LLM while being compatible with PSC detention problem settings. For the supervised fine-tuning (SFT) phase of the LLM, we redefine the detention prediction task as a ranking and strategic decision-making problem, rather than a binary classification problem. The LLM can learn to assign a continuous confidence score, interpreted as the detention probability, to each sample, facilitating more accurate decision-making. Inspired by the Chain of Thought(CoT), this grouping and ranking strategy encourages LLM to think and respond progressively and reasonably while mitigating challenges posed by imbalanced data distributions. Finally, in the inference phase, the LLM computes a detention probability score for each individual sample, allowing users to manually set dynamic thresholds to determine the proportion of ships subject to detention.

Our MSD-LLM not only introduces a novel approach to accurately learning PSC data representations but also marks a breakthrough in combining port risk prediction with LLM for real-world maritime applications. By leveraging PSC domain knowledge, the dual-RSRAE representation learning module, and the robust predictive capabilities of LLMs, this framework paves the way for further advancements in ship risk prediction. Extensive experiments conducted on 31,707 real-world PSC inspection records from the Asia-Pacific region demonstrate the superior performance of MSD-LLM in detention prediction. Our model achieves an average precision improvement of over 8\% on average precision and 13\% on average recall, surpassing state-of-the-art methods and proving its effectiveness in handling real-world maritime risk assessment challenges. From a practical perspective, the MSD-LLM enables accurate prediction of ship detentions under imbalanced PSC data, enhancing the feasibility of maritime safety enforcement. Our method helps alleviate maritime risks, preserve the marine environment, and ensure safe and equitable living and working conditions for crew members as well as enhancing port efficiency. As a result, our approach not only strengthens regulatory compliance but also enhances the overall efficiency of maritime logistics.

\section{Related Work}
\subsection{Ship Detention for PSC}
Ship detention classification or prediction using PSC data has long been a challenging problem. Several studies have focused on ship detention tasks, leading to various classification models ~\cite{79yan2021artificial,90wu2022ship,111yan2022ship,119tian2022cost,132yang2024data,133yang2023machine,145xiao2024interpreting}. Among that, Bayesian Networks (BNs) have been widely used to estimate detention probabilities \cite{132yang2024data,133yang2023machine}. The Key challenge that all those models suffered is the severe class imbalance in PSC datasets, prompting specialized and various strategies to address it.

For instance, Yan et al.\cite{79yan2021artificial} developed a Balanced Random Forest (BRF) model to alleviate the imbalanced distribution problem by adapting the traditional Random Forest (RF) with a balanced structure, ensuring that detained and non-detained samples are more evenly represented during training. Meanwhile, Xiao et al.\cite{145xiao2024interpreting} proposed a novel RF-based model that addresses both data imbalance and prediction uncertainty in detention prediction by identifying key influencing factors that contribute to a vessel being detained. 

In addition, Tian et al.\cite{119tian2022cost} tackled the imbalanced and unlabeled nature of ship inspection records in PSC—due to the large number of uninspected ships—by developing a cost-sensitive learning and semi-supervised learning model. Their method extends the traditional logistic regression approach by introducing a cost parameter designed to account for these specific issues in the data. Furthermore, Yan et al.\cite{111yan2022ship} treated ship detention as an anomaly event, based on the observation that detention is a rare outcome associated with distinct input features when compared to ships that are not detained.

From the above review, it can be seen that artificial intelligence models for ship detention prediction in PSC inspections have primarily relied on ML techniques, and few DL models or structures are specifically designed for the ship detention task. Moreover, the imbalanced data distribution of PSC significantly impacts the performance of the ship detention task.

\subsection{Large Language Models for AI applications}
Recent advancements in LLMs like GPT \cite{OpenAI2023GPT-4}, LLaMa \cite{LLaMa1,LLaMa2,LLaMa3}, and Qwen \cite{Qwen2023,Qwen2024} have greatly expanded their capabilities in grounding, reasoning and action predicting. Beyond text-based inputs, more modal input are involved in current structure to adapt more multi-modal applications. Thus, integrating features from M-LLMs has become a growing area of active research. M-LLMs typically consist of three core components: an encoder for non-text modalities, an LLM as the base model, and a connector to align non-text data with text. These models bridge textual data with inputs like images, videos, audio, and structured data, enabling a range of applications such as intelligent agents \cite{hong2023cogagent,wang2025mobileagenteselfevolvingmobileassistant}, autonomous driving \cite{yuan2024rag}, image generation \cite{qin2024diffusiongpt}, and video understanding \cite{tang2023video} and other domain specific LLM applications\cite{ouyang2024trafficgpt}. 

While LLMs and M-LLMs have been explored in traffic management \cite{ouyang2024trafficgpt} and mobility systems \cite{liang2024exploring}, their application in the maritime domain remains limited as discussed in \cite{wang2024sora}. Inspired by these advancements, we propose MSD-LLM, a multi-modal LLM for ship detention prediction. Our model integrates dual subspace recovery (DSR) features from PSC data with a grouping and ranking strategy, effectively aggregating feature representations while leveraging the predictive power of large language models.

\section{Auto-encoder Representation Learning}
\begin{figure*}[t]
    \centering
    \includegraphics[width=0.78\linewidth]{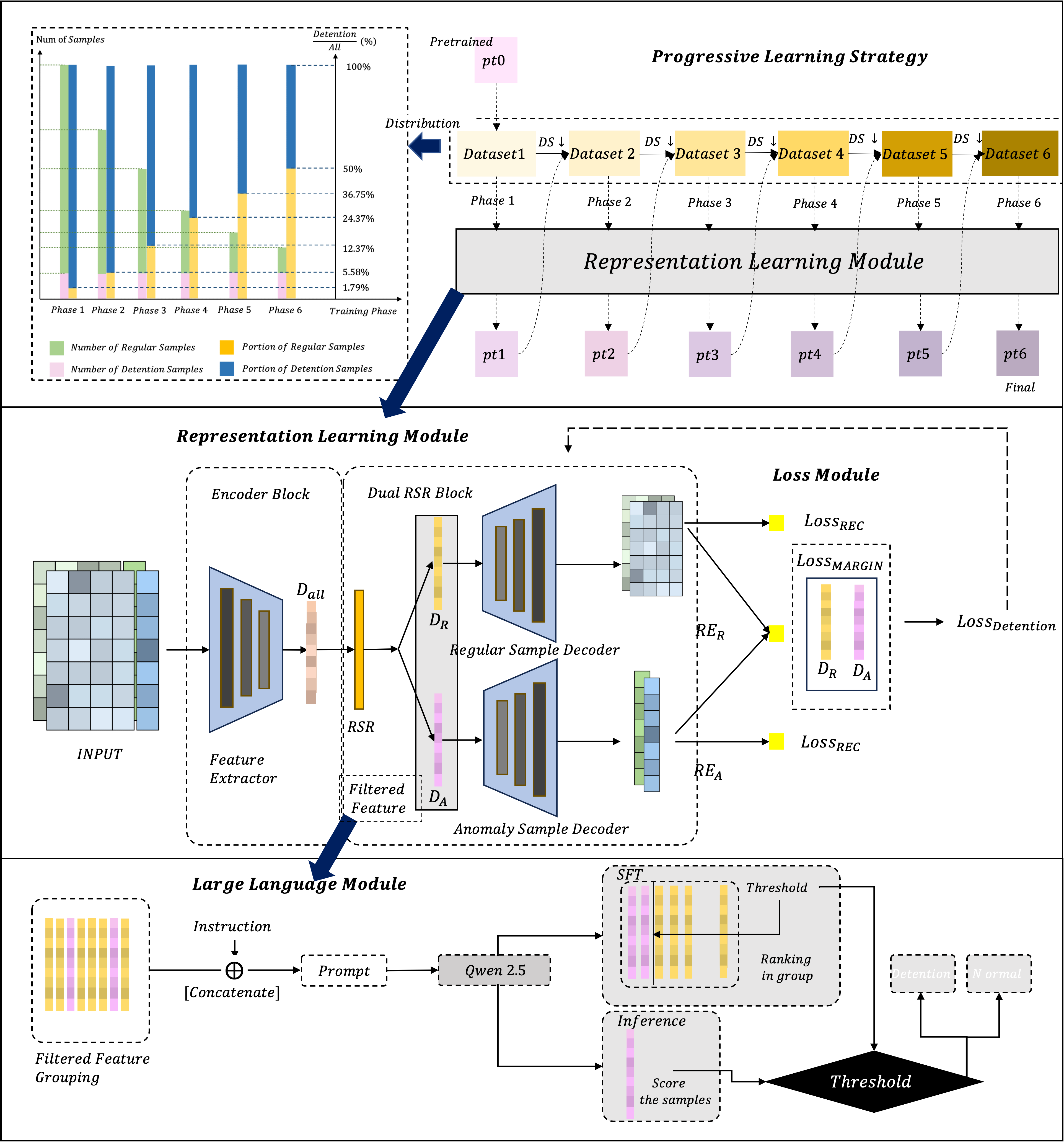}
    \caption{Overview of MSD-LLM, including progressive learning strategy(top sub-figure), the DSR representation learning module(middle sub-figure) and LLM prediction module(bottom sub-figure).}
    \label{fig: overview}
    \vspace{-1em}
\end{figure*}
We propose a novel DSR layer-based auto-encoder architecture for ship detention prediction, combining auto-encoder to extract PSC representations with a progressive learning strategy to address imbalanced data distribution. This approach improves feature representation and enhances the ability detention prediction. The top and middle sub-figures in Figure \ref{fig: overview} illustrate the overall framework and key components of our method.

\subsection{Feature Semantic Clustering}
\label{fsc}
For PSC data, the attributes of each record exhibit significant variability. Traditional data processing methods treat all attributes as independent factors, overlooking the semantic relationships among them. However, certain attributes share inherent connections that should be considered in analysis. 

We designed a semantic feature clustering method for the proposed DRS layer-based auto-encoder framework, categorizing features into two distinct dimensions to enhance representation learning. The first dimension represents basic vessel particulars, primarily consisting of discrete attributes such as \textit{Ship Tonnage, Flag Performance, Recognized Organization, Company Performance, Classification Society Number, Shape Type, and Ship Age}. The second dimension focuses on historical ship detention records, capturing temporal correlations through attributes such as \textit{Last Deficiency Number, Interval Days, Last Inspection State, the Average Number of Deficiencies in the Last 36 Months, the Max Number of Deficiencies in the Last 36 Months, the Probability of Deficiencies in the Last 36 Months, the Total Number of Deficiencies in the Last 36 Months}. 

By clustering features in this way, we provide a feasible solution for discrete PSC data to be compatible with the input of convolutional nerual networks (CNNs). And during training, the CNNs assign and update differentiated weights to each dimension dynamically, optimizing the PSC data contribution to DSR layer-based representation learning and improving overall predictive performance.

\subsection{DSR layer-based auto-encoder module}
\label{dual}
In this module, we first used a feature encoder to extract features into latent codes. Then a regular sample decoder is utilized for reconstructing normal samples, while a detention feature decoder is utilized for reconstructing detained vessels.

\textbf{\textit{Feature Extractor} ($\mathscr{E}$):}
\label{fe}
In the feature extractor $\mathscr{E}$, after transforming the discrete input data to semantic vectors via feature clustering, we adopt CNNs to capture relationships between each attribute and its adjacent attributes, effectively learning local feature dependencies. Given the input data $\{x^{(t)}\}_{t=1}^{N} \in \mathbb{R}^{M}$, where $X$ represents the data matrix and $x^{(t)}$ denotes $t$-th column of $X$, and the extracted feature representation is computed as:
\begin{equation}
    {d}_{all}^{(t)} = \mathscr{E}(x^{(t)})\in\mathbb{R^D}.
\end{equation}
The overall dimensionality of the extracted feature matrix is given by:
\begin{equation}
    {D}_{all} = \mathscr{E}(X).
\end{equation}

\textbf{\textit{DSR Layer} ($A$):}
\label{rsr}
The Dual Robust Subspace Recovery (DSR) layer acts as a feature-level filter, distinguishing regular samples from anomaly (detention) samples. Once the latent feature representation 
$d_{all}^{(t)}$ is obtained from the feature extractor, the DSR layer is applied to amplify the separation between regular and detention samples. By refining these representations, the DSR layer enhances the model’s accuracy in identifying high-risk ships while maintaining robustness against imbalanced data distributions. The DSR layer can be viewed as a linear transformation represented by 
$A_{reg}, A_{det} \in \mathbb{R}^{( D \times d)}$, which reduces the dimensionality from $D$ to $d$. The transformation is expressed as follows:
\begin{equation}
    \widetilde{d}_{reg}^{(t)} = A_{reg}{d}_{all}^{(t)}\in\mathbb{R}^d.
\end{equation}
Similarly, for the detention branch, we also apply the DSR layer to filter features, enabling separation at the latent feature level. The transformation for the detention branch is represented by the following equation:
\begin{equation}
    \widetilde{d}_{det}^{(t)} = A_{det}{d}_{all}^{(t)}\in\mathbb{R}^d.
\end{equation}
DSR layer on the whole dimension of the latent feature can be calculated by Equation:
\begin{equation}
    \widetilde{D}_{reg} = A_{reg}{D}_{all}.
\end{equation}

Similarly, 
\begin{equation}
    \widetilde{D}_{det} = A_{det}{D}_{all}.
\end{equation}

\textbf{\textit{Regular Sample Decoder:}}
\label{rsd}
Traditional RSR-based auto-encoder\cite{lai2019robust} for anomaly detection, adopted only one decoder to reconstruct the regular samples as shown in Equation:  
\begin{equation}
    \widetilde{x}_{reg}^{(t)} = \mathscr{D}(\widetilde{d}_{reg}^{(t)})\in\mathbb{R}^M.
\end{equation}
And the Regular decoder on the whole RSR feature can be described as: 
\begin{equation}
    \widetilde{X}_{reg}= \mathscr{D}(\widetilde{D}_{reg}).
\end{equation}

In our proposed DSR pipeline, we keep this Regular Sample Decoder, using several graph de-convolutional networks to recover the original data information from latent code ${D_{all}}$. In this branch, $RE_R$ is retrieved in a matrix manner. Then, given new data for inference, the new $RE_R$ will determine the possibility of being a regular sample when calculating the distance with $RE_A$.

\textbf{\textit{Detention Feature Decoder:}}
\label{afd}
In addition to the regular sample decoder, we introduce an anomaly feature decoder, supervised specifically by negative (detention) samples. The process begins by applying the RSR layer to filter anomaly samples. Given the highly imbalanced distribution of detention samples in practical PSC data, it is essential to preserve the anomaly representation for effective training in detention prediction tasks.
To achieve this, the RSR layer filters anomaly samples in parallel with regular samples. The transformation for anomaly samples is represented as:
\begin{equation}
    \widetilde{x}_{det}^{(t)} = \mathscr{D}(\widetilde{d}_{det}^{(t)})\in\mathbb{R}^M.
\end{equation}
And the Anomaly decoder on the whole DSR feature can be described as: 
\begin{equation}
    \widetilde{X}_{det}= \mathscr{D}(\widetilde{D}_{det}).
\end{equation}

\textbf{\textit{Detention Loss Function:}}
\label{loss1}
During the training process for detention prediction tasks, we use two types of loss functions to achieve two primary objectives: minimizing reconstruction error and maximizing the separation between regular and anomaly samples. This dual-objective approach ensures the model accurately reconstructs input data while enhancing its discrimination ability. 

The reconstruction loss is computed by minimizing a combined loss function, which consists of the autoencoder reconstruction loss and the DSR layer loss. For $p > 0$, the 
 $l_{2,p}$  reconstruction loss for the autoencoder is defined as:

\begin{equation}
    \mathcal{L}_{reg}^{p}(\mathscr{E}, A, \mathscr{D}_{reg}) = \sum_{t=1}^N ||x_{reg}^{(t)} - \widetilde{x}_{reg}^{t}||_{2}^p.
\end{equation}
Similarly for anomaly branch, 
\begin{equation}
    \mathcal{L}_{det}^{p}(\mathscr{E}, A, \mathscr{D}_{det}) = \sum_{t=1}^N ||x_{det}^{(t)} - \widetilde{x}_{det}^{t}||_{2}^p.
\end{equation}
Original RSR based auto-encoder needs to recover a linear subspace, or equivalently, an orthogonal projection $P$ onto this subspace. Given a dataset $\{y^{(t)}\}^N_{t=1}$ and let $I$ denote the identity matrix in the ambient space of the dataset. The goal is to find an orthogonal projector $P$ of dimension $d$ that robustly approximates the dataset. The least $q^{th}$ power deviations formulation for $q> 0$ minimizes the following loss:

\begin{equation}
    \hat{\mathcal{L}}(P) = \sum^N_{t=1}||(I-P) y^{(t)}||^q_2.
\label{eq:sum}
\end{equation}

The solution of this problem is robust to some outliers when $q 
\leq 1$; In order to void local minima in DSR scenario, we choose $q = 1$ for the final loss function: 
\begin{equation}
\begin{aligned}
    \mathcal{L}^q_{DSR} &= \lambda_1\mathcal{L}_{DSR_1}(A) + \lambda_2\mathcal{L}_{DSR_2}(A) \\
    &:=\lambda_1 \sum^N_{t=1}||d^{(t)}-A^T(\underbrace{Ad^{(t)}} _ {\widetilde{d}^{(t)}})||^q_2 + \lambda_2||AA^T - I_d ||^2_F,
\label{eq:add}
\end{aligned}
\end{equation}
where $A^T$ denotes the transpose of A, Id denotes the $d \times d$ identity matrix and $||\cdot ||_F$ denotes the Frobenius norm. Here $\lambda_1, \lambda_2> 0$ control the balance between terms.
In addition, we introduce a margin loss using SimCSE to enlarge the separation between regular and detention samples, improving class discrimination. Minimizing this loss maximizes the margin between the two classes, enhancing the model’s ability to distinguish them. The margin loss is defined as:
\begin{equation}
        \mathcal{L} = -\log \frac{\exp(\text{sim}(x^{(t)}\tau)}{\sum_{G^- \in \mathcal{N}} \exp(\text{sim}(\mathscr{F}_\mathcal{G}(G_{gt}), \mathscr{F}_\mathcal{G}(G^-)) / \tau)},
\end{equation}
where $sim$ denotes the similarity calculations; here, we use cos similarity. 
Finally, we add the DSR reconstruction loss and Margin loss together with settled weights to get the final loss. 
\begin{table*}[ht]
    \centering
    \caption{Quantitative results of different state-of-the-art ship detention prediction methods or models on global overseas ports and Singapore ports in Tokyo MoU dataset respectively  }
    \begin{tabular}{ccccc|cccc}
    \hline
        &\multicolumn{4}{c}{Global Overseas Ports} & \multicolumn{4}{c}{Singapore Ports}\\
        \cline{2-5}
        \cline{6-9}
        & Precision& Recall & F-score & AUC & Precision& Recall & F-score & AUC \\ \hline
         RF & 0.52 & 0.50 & 0.51 & 0.50 & 0.50 & 0.49 & 0.49 &  0.51\\
         BRF& 0.61 & 0.85 & 0.64 & 0.85 & 0.58 & 0.74 & 0.65& 0.72\\ 
         GBDT & 0.68 & 0.56 & 0.58 & 0.56 & 0.63 & 0.59 &0.61 &0.60\\ 
         iForest & 0.59 & 0.71 & 0.62 & 0.71& 0.55 & 0.72 & 0.62& 0.61\\ \hline
         Plain-AE &0.58&0.72&0.62&0.72  & 0.51& 0.64 &0.59 &0.63\\ 
         RSRAE & 0.52 &0.73 & 0.61&0.72 & 0.49& 0.66&0.56& 0.74\\ 
         DSRAE(Ours)& 0.65 & 0.86& 0.74&0.87&0.61& 0.77&0.68&0.77  \\ \hline
         Qwen-2.5(base) & 0.58&0.72 & 0.64& 0.75 &0.59&0.71&0.64&0.70\\
         MSD-LLM &\textbf{0.70}&\textbf{0.88}& \textbf{0.77}&\textbf{0.90} &\textbf{0.72}&\textbf{0.85}&\textbf{0.77}& \textbf{0.89} \\ \hline
         
    \end{tabular}

    \label{tab:detention}
\end{table*}

\subsection{Progressive Learning}
We employ progressive learning to address the challenge of extremely imbalanced data distribution in PSC detention tasks. In real-world scenarios, the distribution of detention samples can vary significantly between training data and realistic testing or inference datasets, making adaptability crucial for reliable predictions.

To enhance generalization, we progressively adjust the training dataset to reflect different real-world distributions. As shown in top sub-figure in Figure \ref{fig: overview}, the training process is divided into six phases, with detention sample proportions set at 1.79\%, 5.58\%, 12.37\%, 24.37\%, 36.75\%, and 50\%. We achieve this by reducing the number of regular samples to match the target distribution, allowing the model to adapt more effectively to varying data conditions. The progressive learning pipeline follows a structured stage-wise training approach to incrementally adapting the model to varying data distributions. The process begins with a pre-trained model $pt_0$, which is trained using the phase 1 training dataset. Subsequently, $pt_1$ serves as the pre-trained model for phase 2 training, where it is further refined using the phase 2 dataset. This iterative process continues across all six phases, with each checkpoint progressively adapting to increasingly complex data distributions. By the end of training, the final model checkpoint captures a robust and well-generalized representation, enabling it to handle diverse real-world PSC data distributions more effectively.

\section{LLM Aggregation}
\subsection{Problem Setting and Prompt Engineering}
Our proposed MSD-LLM deviates from the traditional ship detention classification paradigm by assigning a continuous confidence score to each sample, reflecting its detention probability. Rather than directly regressing the score, we utilize the reasoning capabilities of the LLM through a group-based ranking strategy. The details of our MSD-LLM are shown in the bottom sub-figure of Figure \ref{fig: overview}. First, vessel samples are clustered into groups of 20, with each group containing two detention samples and the rest as regular samples. Then, the LLM predicts the confidence score (detention probabilities) of each samples in a group and ranks them. This design enables the model to learn ranking patterns instead of making isolated predictions. The grouping strategy is adaptable to different dataset characteristics, helping to mitigate data imbalance and enhance prediction robustness.

For prompt engineering, in our MSD-LLM framework, we integrate the latent representation extracted from the Dual-RSR layer with an LLM to predict detention probability scores. To fully utilize the LLM’s reasoning capabilities, we structure the input prompt similarly to M-LLMs. Specifically, the representation features are concatenated with an instruction token, enabling the LLM to process and interpret structured embeddings within a natural language reasoning framework. This design allows the model to combine DSR’s powerful feature extraction with the contextual understanding and decision-making abilities of LLMs, resulting in more accurate and interpretable detention predictions.

\subsection{Supervised Finetuning}
The detailed pipeline is shown in Figure \ref{fig: overview}. During the SFT phase, we trained Qwen-VL-2.5\cite{Qwen2024}, replacing its original ViT feature extractor with our DSR feature representation. The input consists of groups of 20 DSR features, with the large language model (LLM) tasked with ranking these features based on the likelihood of a ship being detained. Each feature within the group is assigned a floating-point score representing the detention probability.

During inference, the LLM evaluates newly provided features and assigns a detention probability score, enabling a flexible decision-making process. This ranking-based approach, rather than direct classification, improves interpretability and robustness in handling imbalanced detention datasets. It also allows the LLM to generalize across varying distributions, ensuring high predictive accuracy in maritime risk assessment.

\section{Experiment}
\subsection{Dataset and Experiment Settings}
Our study utilizes 317,170 Port State Control (PSC) records from the Tokyo MoU, partitioned into training (70\%), validation (20\%), and test (10\%) sets. For evaluation, we further divide the test data into two perspectives: global overseas data, containing inspection records from multiple ports worldwide expect Singapore ports, and regional port data, using Singapore only. These datasets differ significantly in size and distribution—global data aligns with the training distribution, while Singapore data follows a different pattern. This setup enables a comprehensive robustness assessment under varied real-world conditions.

For model training, we use two Nvidia A100 GPUs for DSR-layer based auto-encoder and eight Nvidia A100 GPUs for the large language model (LLM). The pipeline is trained with a learning rate of $10^{-4}$ for the first 15 epochs, followed by $10^{-5}$ for 10 epochs, and $10^{-6}$ for the final 5 epochs. After 30 epochs, the training loss stabilizes, ensuring model convergence and robustness.

\subsection{Quantitative Results for Detention}
We compared our proposed DSRAE with other architectures in detention prediction tasks, including traditional ML model such as RF\cite{breiman2001random}, BRF\cite{79yan2021artificial}, GBDT\cite{breiman1997arcing}, iForest\cite{4781136}, auto-encoder based DL model such as plain-AE \cite{ballard1987modular}, RSRAE\cite{lai2019robust}, our representation module DSRAE, and large language model such as Qwen2.5\cite{Qwen2024} base model and our proposed MSD-LLM. We evaluate those models using average precision, recall. F-score and AUC. 
The results are shown in table \ref{tab:detention}.

According to Table \ref{tab:detention}, the results presented demonstrate the superior performance of MSD-LLM compared to other models in ship detention prediction across both Tokyo MoU and Hifleet datasets. The performance boost highlights the effectiveness of our dual-subspace representation learning strategy (DSRAE) and the LLM-based grouping and ranking strategy, which collectively enhance prediction accuracy. Notably, DSRAE alone already improves upon traditional methods, but the integration with MSD-LLM further enhances the overall performance, demonstrating its ability to handle imbalanced data distributions and capture intricate detention patterns more effectively.

\subsection{Ablation Study}
\textbf{\textit{Progressive Learning:}}
\begin{figure*}
    \centering
    \includegraphics[width=1\linewidth]{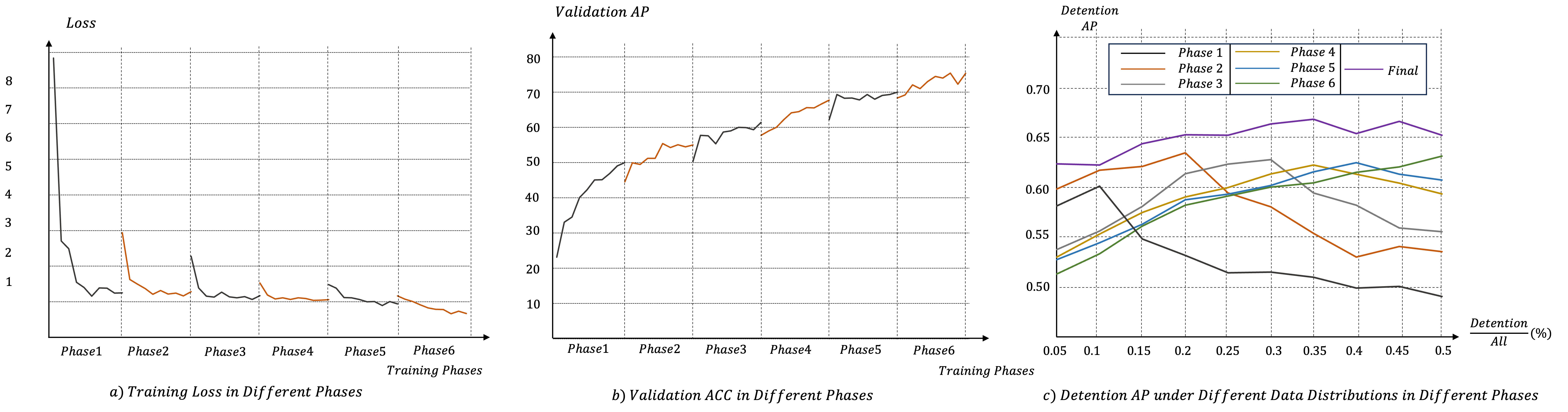}
    \caption{Training Loss(a) and Validation average precision(b) of different training phases in progressive training strategy. (c)demonstrates the impact of training data distribution on the final results. (c) also shows that aggregating different phases together can make the model compatible with different data distributions.}
    \label{fig:pro}  
    \vspace{-1em}
\end{figure*}
The ship detention prediction task faces a highly imbalanced data distribution, making accurate classification challenging. This issue is compounded in real-world scenarios, where the distribution of evaluation datasets is often unknown and difficult to predict. To address this, we propose a progressive learning pipeline that strategically downsamples regular samples, improving adaptability to various distributions of regular and detained ships. This approach enhances the model’s robustness and generalization, ensuring its effectiveness for practical maritime risk assessment.

Figure \ref{fig:pro} illustrates the differences between regular training and progressive learning. Sub-figures (a) and (b) show the training loss and validation average precision during progressive learning. At each phase transition, there is a temporary surge in loss and a drop in precision, caused by the model adjusting to a new data distribution after potential overfitting in the previous phase. Despite these fluctuations, the overall trend reveals a steady reduction in loss and consistent improvement in precision, demonstrating the strategy’s effectiveness.

Additionally, sub-figure (c) highlights the impact of different test data distributions. Models trained on individual phases perform well on their specific distribution but decline on others. In contrast, the full progressive learning strategy yields a more robust model, with smoother performance and higher overall accuracy across various distributions.
\begin{table}[t]
    \centering
     \caption{Comparison between straightforward attribute clustering with our proposed feature clustering}
    \begin{tabular}{ccc}
    \hline
         & Precision& AUC \\ \hline
         RSR-based AE(Plain) &0.4072 &0.6641 \\ 
         DSR-based AE(Plain)&0.6504 &0.7612 \\
         RSR-based AE(Final) &0.5238&0.7179\\
         DSR-based AE(Final)&\textbf{0.6518} &\textbf{0.8722}\\ \hline
    \end{tabular}
   \vspace{-1em}
    
    \label{tab:cpm}
\end{table}

\textbf{\textit{RSR vs DSR:}}
\label{dvm}
The traditional RSRAE structure \cite{lai2019robust} was originally developed for anomaly detection. However, our experiments revealed that RSR struggles with highly imbalanced data distributions, such as those in ship detention prediction tasks. To overcome this limitation, we developed DSRAE, which adds a dedicated branch for anomaly recovery, enhancing the model’s ability to learn sparse anomaly samples and improving both robustness and generalizability.

\begin{figure}[t]
    \centering
    \includegraphics[width=0.85\linewidth]{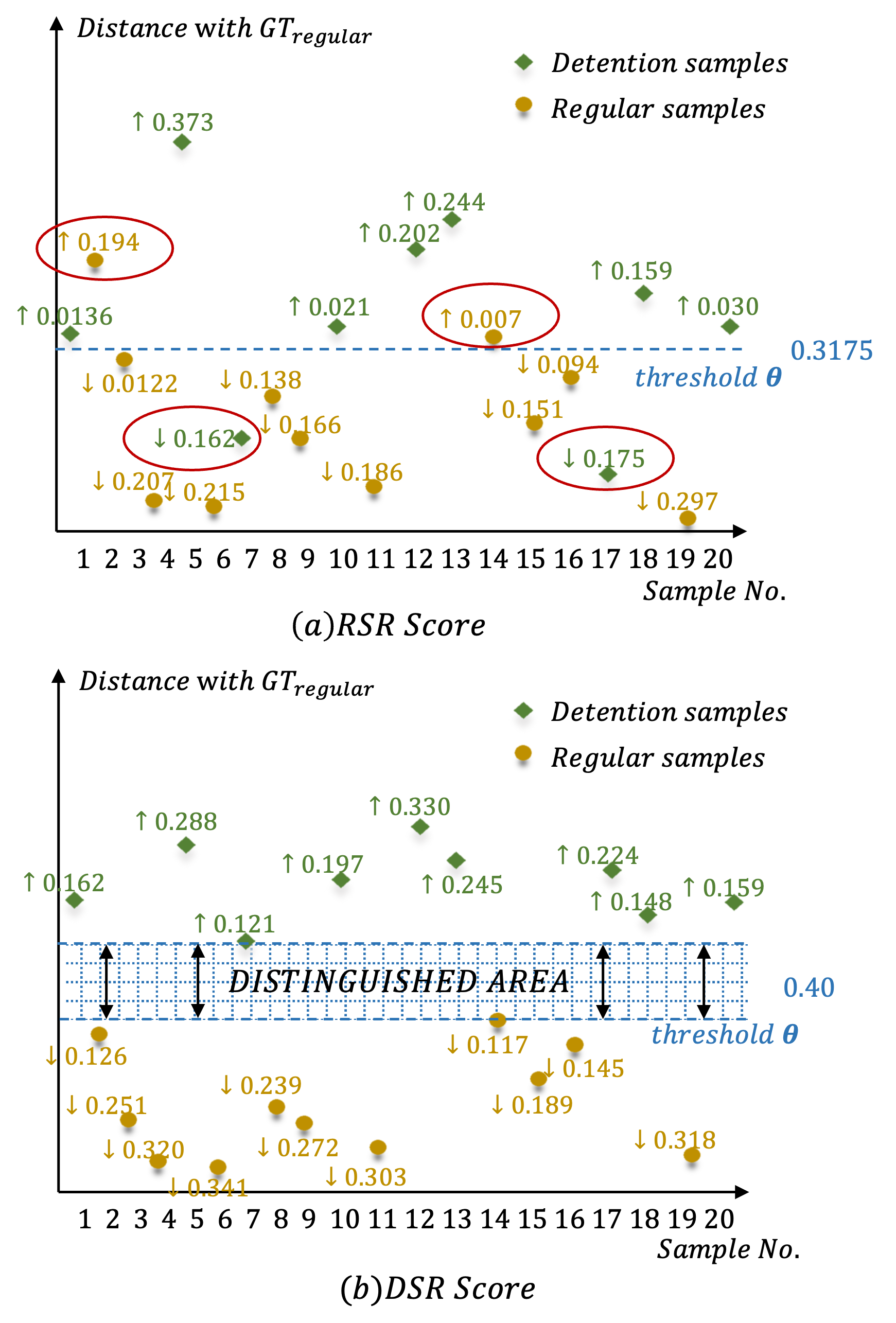}
    \caption{ Visualization of sample distance to the threshold. (a)shows the score of RSR-layer based representation learning module while (b) indicates the score of DSR-layer based module.}
    \label{fig:training Gap}
    \vspace{-1em}
\end{figure}

As shown in Table \ref{tab:cpm}, we compared plain RSR and DSR layer based auto-encoder (without optimization techniques like progressive learning or a deficiency branch). DSR achieved a 0.25 improvement in Average Precision (AP) and 0.10 in AUC, demonstrating its superior capability to capture detention-related features through its dual-decoder structure. In a comprehensive comparison of full RSR and DSR pipeline with all optimizations applied, including progressive learning, the DSR consistently outperformed the RSR-based version, reinforcing the effectiveness of our approach. Additionally, Figure \ref{fig:training Gap} shows that during training, DSRAE is more effective at enlarging the gap between regular and detention samples, improving class separation, and boosting overall predictive performance.
\begin{table}[ht]
    \centering
    \caption{The impact on different parameters choice of group strategy.}
    \begin{tabular}{ccc }
    \hline
         & AP & Recall \\
         \hline
        5 per group & 0.65&  0.86\\
        12 per group &0.67 & \textbf{0.88} \\
        20 per group & \textbf{0.70} & \textbf{0.88}\\
        50 per group & 0.52&  0.68 \\
        \hline
    \end{tabular}
    \vspace{-1em}
    \label{tab:group}
\end{table}
\textbf{\textit{Grouping Strategy Parameters:}}
During the Supervised Fine-Tuning (SFT) phase of the LLM, our proposed MSD-LLM clusters the DSR layer-based representations into distinct groups, ensuring each group contains a fixed number of regular and detained samples. Determining the optimal group size is crucial, as it significantly impacts model performance.

If the group size is too small, training becomes time-consuming due to the large number of iterations required, and group diversity decreases, limiting the LLM’s ability to learn an effective ranking strategy. Conversely, a large group size increases computational costs, slows convergence, and reduces ranking precision, making it harder to distinguish high-risk ships from regular ones. After extensive experimentation, we found that a group size of 20 achieves the best performance, as shown in Table \ref{tab:group}.

\textbf{\textit{The Effectiveness of LLM and Group Strategy:}}
Integrating LLMs with PSC data representation is challenging, but our findings show that accurate ship detention prediction is only possible when LLMs are combined with our grouping strategy. Unlike conventional autoencoder-based learning or direct LLM integration, MSD-LLM reframes detention prediction as a reasoning task involving sample ranking and decision-making.

To validate our approach, we compare several model configurations. We first evaluate RSR and DSR layer-based auto-encoders as baselines and then test LLM integration with and without the grouping strategy. As shown in Table \ref{tab:LLM}, combining LLMs with the grouping strategy (GS-LLM) consistently achieves the best performance across all settings. In contrast, direct LLM concatenation without the grouping strategy fails to improve accuracy and can even degrade performance despite increasing model parameters. These results confirm that while LLMs can enhance ship detention prediction, their success depends on adopting an effective strategy such as grouping.
\begin{table}[]
    \centering
    \caption{The effectiveness of LLM and Grouping Strategy(GS)}
    \begin{tabular}{cccc}
    \hline
    &AP & Recall &AUC\\ \hline
    RSR-based autoecoder &0.5238&0.7302&0.7179 \\
    RSR + LLM(Plain) &0.5351&0.6977& 0.6645 \\
    RSR + GS-LLM(GS)&0.6012&0.7718&0.8154 \\ \hline
    DSR-base autoencoder  & 0.6518&0.8616& 0.8722\\
    DSR + LLM (Plain) &0.5775&0.7002&0.7043 \\ 
    DSR + GS-LLM(MSD-LLM)   & \textbf{0.7012}&\textbf{0.8751} &\textbf{0.8969}\\
         \hline
    \end{tabular}
    \vspace{-1em}
    \label{tab:LLM}
\end{table}

\section{Conclusion}
Ship detention plays a crucial role in maritime port risk management, ensuring regulatory compliance and mitigating hazards. However, traditional classification-based ML models struggle with complex data interdependencies and limited representation learning. While auto-encoder-based DL models enhance feature extraction, they still face challenges with imbalanced data distribution. To address these issues, we propose the MSD-LLM, integrating robust representation learning ability of DSR with the predictive and reasoning capabilities of LLMs. This approach effectively handles diverse data distributions and adapts to varied practical scenarios, significantly improving ship detention prediction accuracy and enhancing port risk assessment and maritime safety enforcement.

MSD-LLM offers a new perspective on integrating LLMs into maritime AI, marking a significant advancement in ship detention prediction. While prior approaches have made progress, they fall short of industrial-level requirements due to scalability, accuracy, and adaptability limitations. By leveraging LLM reasoning capabilities alongside the data distribution insensitivity of our pipeline, MSD-LLM effectively bridges the gap between academic research and real-world applications. Looking ahead, we believe that further integration of domain-specific knowledge with LLMs will drive more efficient and intelligent maritime AI solutions.

\bibliographystyle{IEEEtran}
\bibliography{fig/IEEEtran}

\end{document}